\documentclass[11pt]{article}
\usepackage[hyperref]{ccl2024-en}
\usepackage{times}
\usepackage{url}
\usepackage{latexsym}
\usepackage{fancyhdr}
\usepackage{graphicx}
\usepackage{booktabs}
\usepackage{multirow}

\title{MFE-NER: Multi-feature Fusion Embedding for Chinese Named Entity Recognition}

\author{
Jiatong Li$^1$\footnote{Corresponding Author}\and
Kui Meng$^2$\footnote{Corresponding Author}\\
\affiliations
$^1$The University of Melbourne\\
$^2$Shanghai Jiao Tong University\\
\emails
jiatongl3@student.unimelb.edu.au,
mengkui@sjtu.edu.cn,
}

\author{Jiatong Li \\
  The University of Melbourne \\
  {\tt jiatongl3@student.unimelb.edu.au} \\\And
  Kui Meng \\
  Shanghai Jiao Tong University \\
  {\tt mengkui@sjtu.edu.cn} \\}

\date{}

\begin{document}
\maketitle
\begin{abstract}
In Chinese Named Entity Recognition, character substitution is a complicated linguistic phenomenon. Some Chinese characters are quite similar as they share the same components or have similar pronunciations. People replace characters in a named entity with similar characters to generate a new collocation but referring to the same object. As a result, it always leads to unrecognizable or mislabeling errors in the NER task. In this paper, we propose a lightweight method, MFE-NER, which fuses glyph and phonetic features, to help pre-trained language models handle the character substitution problem in the NER task with limited extra cost. Basically, in the glyph domain, we disassemble Chinese characters into Five-Stroke components to represent structure features. In the phonetic domain, an improved phonetic system is proposed in our work, making it reasonable to describe phonetic similarity among Chinese characters. Experiments demonstrate that our method performs especially well in detecting character substitutions while slightly improving the overall performance of Chinese NER.
\end{abstract}

\section{Introduction}
\begin{figure}[htbp]
  \centering
  \includegraphics[width=0.45\textwidth]{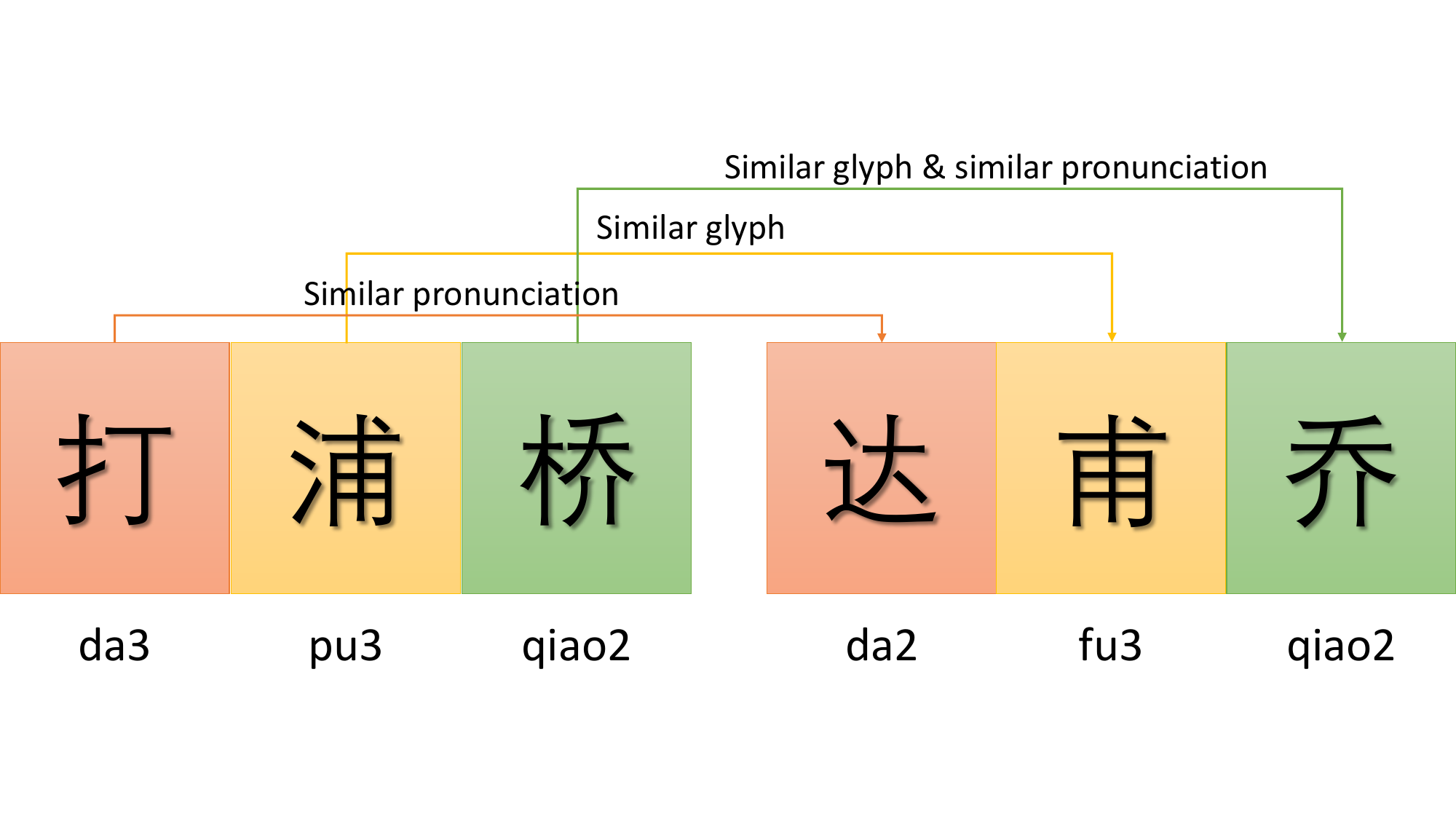}
  \caption{The substitution example of a Chinese word. On the left is a famous place in Shanghai. On the right is a new word after character substitution, which is more like a person or a brand.}
  \label{fig:fig1}
\end{figure}

Recently, pre-trained language models have been widely used in Natural Language Processing (NLP) ~\cite{2021SMedBERT}, constantly refreshing the benchmarks of specific NLP tasks. By applying the transformer structure, semantic features can be extracted more accurately. However, in the Named Entity Recognition (NER) area, tricky problems still exist. Most significantly, the character substitution problem severely affects the performance of NER models. To make things worse, this problem has become even more common these years, especially in social media. Due to the particularity of Chinese characters, there are multiple ways to replace original Chinese characters in a word. Characters with similar meanings, shapes, or pronunciations can be selected for character substitution. A simple example shown in Figure \ref{fig:fig1} is a Chinese word, which represents a famous place in Shanghai. Here, all three characters in the word are substituted by other characters with similar glyphs or pronunciations. After substitution, this word looks more like the name of a person rather than a place.\\
In practice, it is extremely hard for those pre-trained language models to tackle this problem. Currently, the tasks for pre-training Chinese language models are mainly focused on the semantic domain, neglecting glyph and phonetic features. However, most character substitution cases exist in glyph and phonetic domains. At the same time, social media hot topics are changing rapidly, creating new expressions or substitutions for original words every day. It is technically impossible for pre-trained models to include all possible collocations. Models that only saw the original collocations before will naturally suffer from Out-of-Vocabulary (OOV) problems. \\
In this paper, we propose a lightweight method, Multi-feature Fusion Embedding for Chinese Named Entity Recognition (MFE-NER), which fuses extra glyph and phonetic features to detect possible substitution forms of named entities in Chinese. On top of using pre-trained models to represent the semantic feature, we choose a structure-based encoding method, known as 'Five-Strokes', to introduce glyph features. In the phonetic domain, we propose `Trans-pinyin', which combines 'Pinyin', a unique phonetic system, with international standard phonetic symbols. Experiments on three general datasets and our special-designed dataset with character substitutions have illustrated that with limited extra cost, MFE-NER can not only help NER models handle the character substitution problem, but also enhance the overall performance of NER models, which makes it especially suitable to be used in current language environments. \\
To summarize, our major contributions are:
\begin{itemize}
    \item We emphasize the influence of character substitution in the NER task and propose to introduce extra information from Chinese characters to handle this problem.
    \item For glyph features of Chinese characters, we apply the Five-Stroke encoding method, denoting structure patterns of Chinese characters.
    \item To represent phonetic features of Chinese characters, we propose a new method named `Trans-pinyin', to make it possible to evaluate phonetic similarity among Chinese characters.
    \item Experiments show that with limited extra cost, our method is more reliable in recognizing substituted Chinese Named Entities while slightly improving the overall performance of NER models.
\end{itemize}

\section{Related Work}
After the stage of statistical machine learning algorithms, Named Entity Recognition has stepped into the era of deep neural networks. Researchers started to use Recurrent Neural Network (RNN) \cite{hammerton2003named} to recognize named entities in sentences based on character embedding and word embedding, solving the feature engineering problems that traditional statistical methods have. Bidirectional Long Short Term Memory (Bi-LSTM) network \cite{huang2015bidirectional} was first applied in Chinese Named Entity Recognition, which became one of the baseline models. The performance of the Named Entity Recognition task thus gets greatly improved. \\
These years, large-scale pre-trained language models based on Transformer \cite{vaswani2017attention} have shown their superiority in Natural Language Processing tasks. The self-attention mechanism can better capture the long-distance dependency in sentences and the parallel design is suitable for mass computing. Bidirectional Encoder Representations from Transformers (BERT) \cite{kenton2019bert} has achieved great success in many branches of NLP. In the Chinese Named Entity Recognition field, these pre-trained models have greatly improved the recognition performance \cite{cai2019research}.\\
However, the situation is more complicated in real-language environments. The robustness of NER models is not guaranteed by pre-trained language models. Researchers have started to introduce prior knowledge to improve the generalization of NER models. SMedBERT \cite{2021SMedBERT} introduces knowledge graphs to help the model acquire the medical vocabulary explicitly. \\
Meanwhile, in order to relieve character substitution problems and enhance the robustness of NER models, researchers have also paid attention to utilizing glyph and phonetic features of Chinese characters. Jiang Yang and Hongman Wang suggested using the `Four-corner' code, a radical-based encoding method for Chinese characters, to represent glyph features of Chinese characters \cite{yang2021incorporating}, showing the advantage of introducing glyph features in Named Entity Recognition. However, the `Four-corner' code is not that expressive because it only works when Chinese characters have the same radicals. Tzu-Ray Su and Hung-Yi Lee suggested using Convolution Auto Encoder to build a bidirectional injection between images of Chinese characters and pattern vectors \cite{su2017learning}. This method is brilliant but lacks certain supervision, which means that it's hard to explain the concrete meanings of the pattern vectors. Fan Yang, Jianhu Zhang et al introduced `Five-Stroke' in NER models, which clearly improves the performance \cite{10.1007/978-3-319-99495-6_16}. Recently, Zijun Sun, Xiaoya Li et al proposed ChineseBERT \cite{sun2021chinesebert}, fusing glyph and phonetic features to the pre-trained models. Their work is impressive in fusing glyph and phonetic features as the input for pre-trained models, but there still exist some problems. For example, using flatten character images is inefficient for adding huge costs to feature dimensions and enlarging possible negative influence on the model. As Chinese characters have a limited number of components in structures, it will be more convenient and efficient to extract and denote their glyph patterns by certain component encoding methods.

\section{Method}
\begin{figure}
  \centering
  \includegraphics[width=0.45\textwidth]{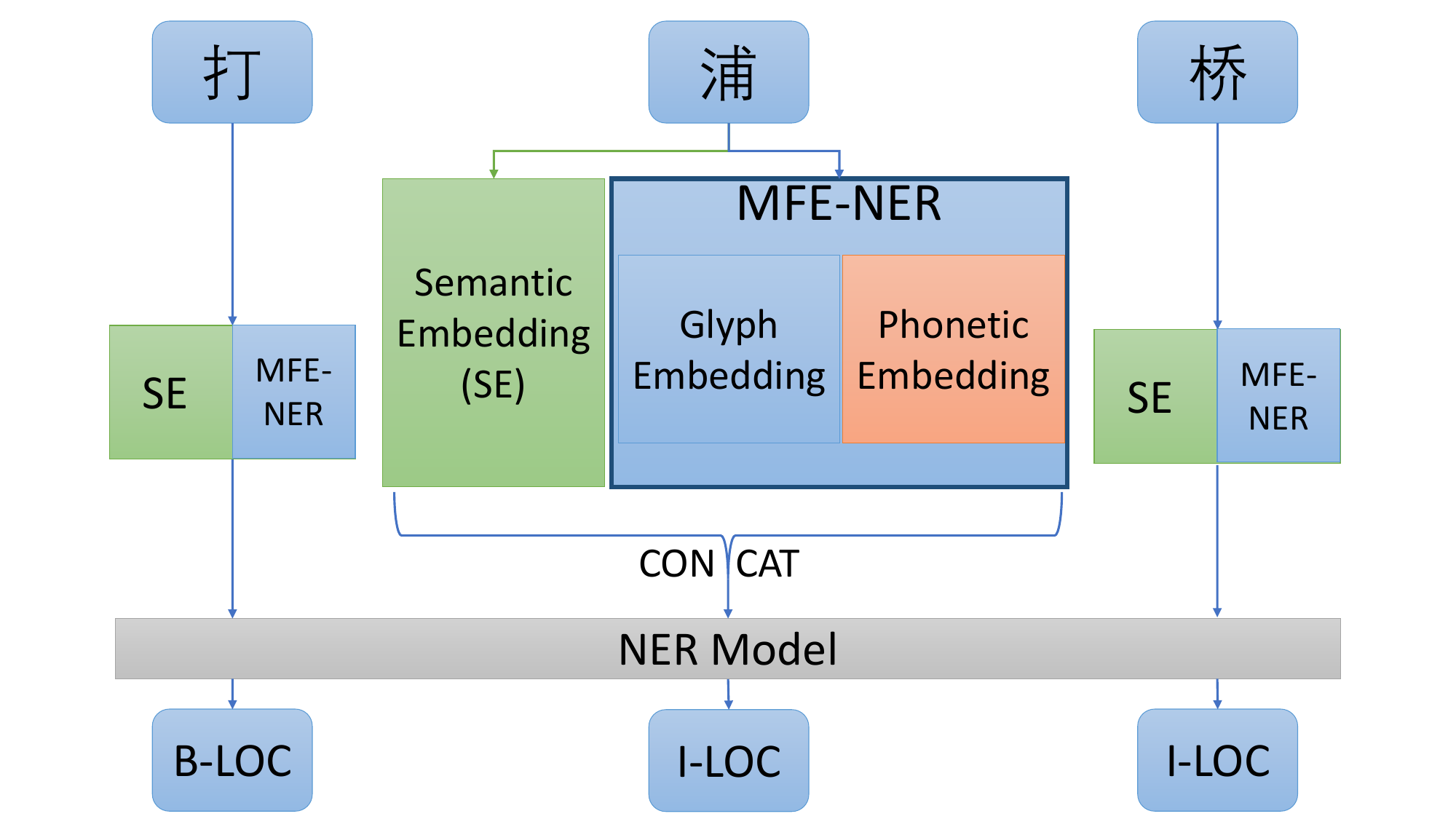}
  \caption{The structure of MFE-NER.}
  \label{fig:fig2}
\end{figure}
our MFE-NER is a lightweight Named Entity Recognition method fusing the glyph and phonetic feature embeddings for Chinese character substitution, which is complementary to pre-trained language models in the representation of Chinese characters. As shown in Figure \ref{fig:fig2}, MFE-NER introduces an extra module, fusing glyph embedding with `Five-Strokes' and phonetic embedding by `Tran-Pinyin'. \\
Basically, for a Chinese sentence $S$ with length $n$, the sentence $S$ is divided naturally to different Chinese characters $S=c_1,c_2,c_3,...,c_n$. Each character $c_i$ will be mapped to an embedding vector $\textbf{e}_i$, which can be divided into the above three parts, $\textbf{e}^s_i$, $\textbf{e}^g_i$ and $\textbf{e}^p_i$. Plus semantic embedding from pre-trained language models, all three embedding parts are designed based on a simple principle, similarity. In this paper, the character similarity between two Chinese characters $c_i$ and $c_j$ is defined by computing their L2 distance in the three aspects. Here, we use $s^s_{ij}$ to denote semantic similarity, $s^g_{ij}$ for glyph similarity and $s^p_{ij}$ for phonetic similarity. So, we have:
\begin{equation}
    s^s_{ij} = \|\mathbf{e}^s_i-\mathbf{e}^s_j\|
\end{equation}
\begin{equation}
    s^g_{ij} = \|\mathbf{e}^g_i-\mathbf{e}^g_j\|
\end{equation}
\begin{equation}
    s^p_{ij} = \|\mathbf{e}^p_i-\mathbf{e}^p_j\|
\end{equation}
In this case, MFE-NER can better represent the inner distribution of Chinese characters in embedding space.

\subsection{Semantic Embedding using Pre-trained Models}
Semantic embedding is vital to Named Entity Recognition. In Chinese sentences, a single character does not mean a word, because there is no natural segmentation in Chinese grammar. So, technically, we have two choices to acquire Chinese embedding. The first way is word embedding, trying to separate Chinese sentences into words and get the vector representation of words, which makes sense but is limited by the accuracy of word segmentation tools. The other is character embedding, which maps Chinese characters to different embedding vectors in semantic space. In practice, it performs better in Named Entity Recognition. Our work also requires to use Chinese character embeddings because glyph and phonetic embeddings are targeted at single Chinese characters. \\
So far, pre-trained models have been widely used in the semantic domain. One typical method is Word2Vec \cite{2013Efficient}, which starts to use static embedding vectors to represent Chinese characters in the semantic domain. Now, we have more options. BERT \cite{kenton2019bert} has its Chinese version and can express semantic features of Chinese characters more accurately, hence having better performance in NER tasks.\\
\subsection{Glyph Embedding with Five-Strokes}
\begin{figure}
  \centering
  \includegraphics[width=0.45\textwidth]{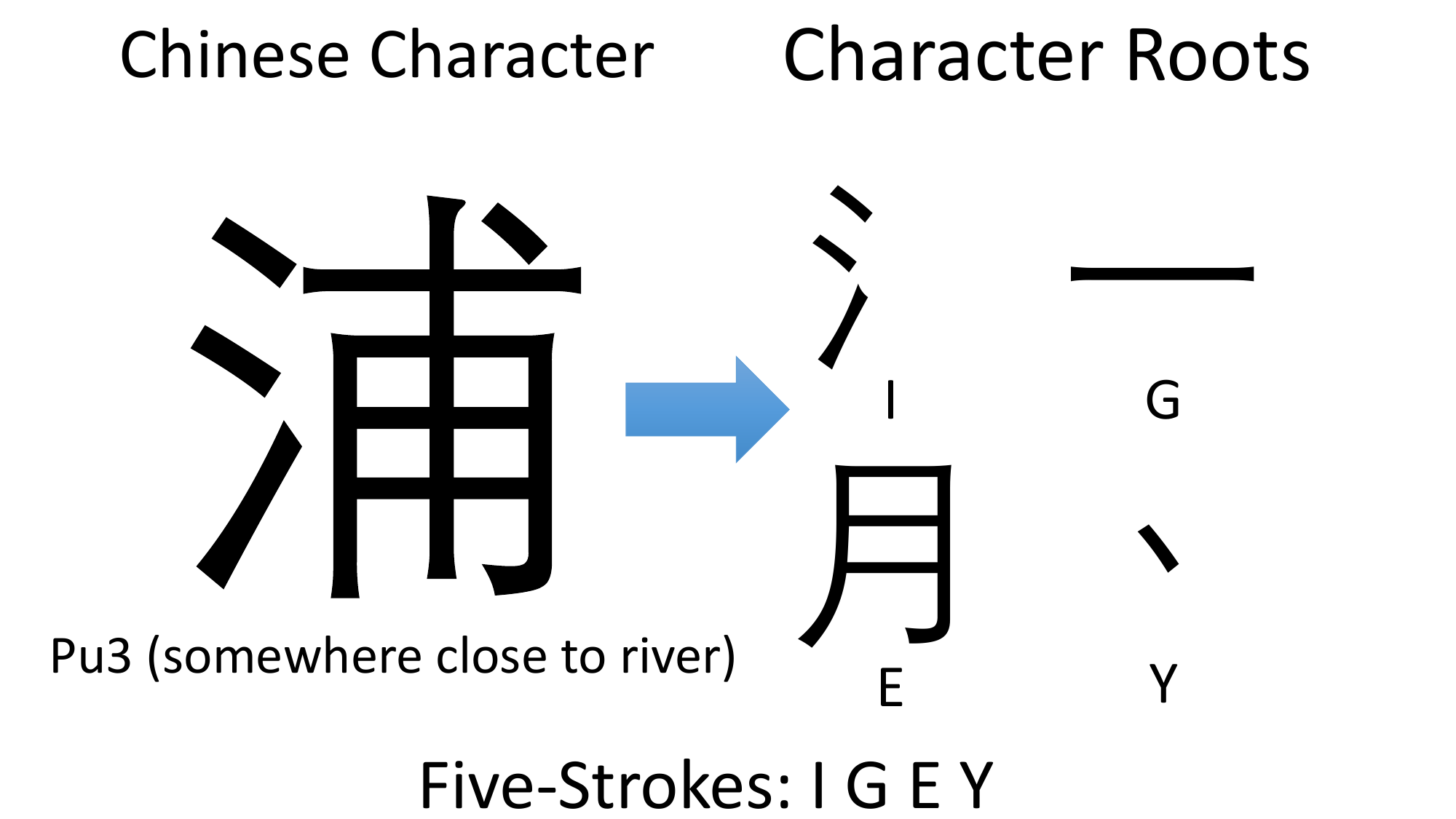}
  \caption{The `Five-Strokes' representation of the Chinese characters, `pu3' (somewhere close to a river). `Five-Strokes' divides the character into four character roots ordered by writing custom so that the structure similarity can be denoted.}
  \label{fig:fig3}
\end{figure}
Chinese characters, different from Latin Characters, are pictographs, which show their meanings in shapes. However, it is extremely hard for people to encode these Chinese characters in computers. A common strategy is to give every Chinese character a unique hexadecimal string, such as `UTF-8' and `GBK'. However, this kind of strategy processes Chinese characters as independent symbols, totally ignoring the structural similarity among Chinese characters. In other words, the closeness in the hexadecimal string value can not represent the similarity in their shapes. Some work has tried to use images of Chinese characters as glyph embedding, which is also unacceptable and ineffective due to the complexity and the large space it will take.\\
In this paper, we propose to use `Five-strokes', a famous structure-based encoding method for Chinese characters, to get our glyph embedding. `Five-Strokes' was put forward by Yongmin Wang in 1983. This special encoding method for Chinese characters is based on their structures. `Five-Strokes' holds the opinion that Chinese characters are made of five basic strokes, horizontal stroke, vertical stroke, left-falling stroke, right-falling stroke and turning stroke. Based on that, it gradually forms a set of character roots, which can be combined to make up the structure of any Chinese character. After simplification for typing, 'Five-Strokes' maps these character roots into 25 English characters (`z' is left out) and each Chinese character is made of at most four corresponding English characters, which makes it easy to acquire and type in computers. It is really expressive that four English characters can have $25^4 = 390625$ arrangements, while we only have about 20 thousand Chinese characters. In other words, `Five-Strokes' has a rather low coincident code rate for Chinese characters.\\
For example, in Figure \ref{fig:fig3}, the Chinese character `pu3' (somewhere close to a river) is divided into four different character roots by Chinese character writing custom, which will later be mapped to English characters so that we can further encode them by introducing one-hot encoding. For each character root, we can get a 25-dimension vector. In this paper, in order to reduce space complexity, we sum up these 25-dimension vectors as the glyph embedding vector. We also list two different characters, `fu3' (an official position) and 'qiao2' (bridge), and calculate the similarity between them. The two characters `pu3' and `fu3' with similar components are close in embedding space, while `qiao2' and `pu3' are much more distant, which gives NER models extra patterns.\\

\subsection{Phonetic Embedding with Trans-pinyin}
\begin{figure}
  \centering
  \includegraphics[width=0.45\textwidth]{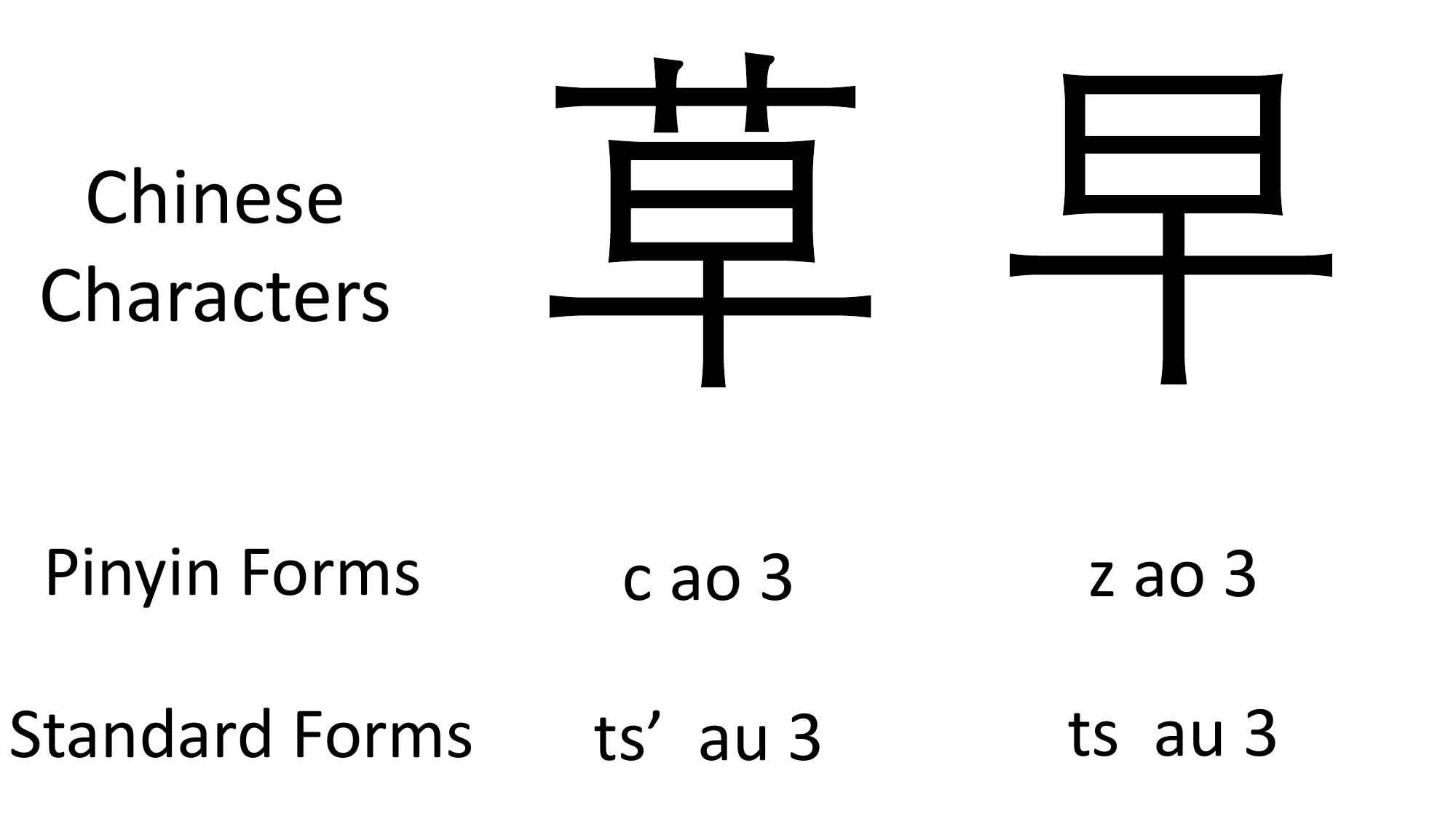}
  \caption{The `Pinyin' form and standard form of two Chinese characters, `cao3' (grass) on the left and `zao3' (early) on the right.}
  \label{fig:fig4}
\end{figure}
Chinese use `Pinyin', a special phonetic system, to represent the pronunciation of Chinese characters. In the phonetic system of `Pinyin', we have four tunes, six single vowels, several plural vowels, and auxiliaries. Every Chinese character has its expression, also known as a syllable, in the `Pinyin' system. A complete syllable is usually made of an auxiliary, a vowel, and a tune. Typically, vowels appear on the right side of a syllable and can exist without auxiliaries, while auxiliaries appear on the left side and must exist with vowels.\\
However, the `Pinyin' system has an important defect. Some similar pronunciations are denoted by totally different phonetic symbols. For the example in Figure \ref{fig:fig4}, the pronunciations of `cao3' (grass) and `zao3' (early) are quite similar because the two auxiliaries `c' and `z' sound almost the same that many native speakers may confuse them. This kind of similarity can not be represented by phonetic symbols in the `Pinyin' system, where `c', and `z' are independent auxiliaries. In this situation, we have to develop a method to combine `Pinyin' with another standard phonetic system, which can better describe characters' phonetic similarities. Here, the international phonetic system seems the best choice, where different symbols have relatively different pronunciations so that people will not confuse them.\\
We propose the 'Trans-pinyin' system to represent character pronunciation, in which auxiliaries and vowels are transformed to standard forms and keep the tune in the `Pinyin' system. After transformation, `c' becomes `$ts$' and `z' becomes `$ts'$', which only differ in phonetic weight. We also make some adjustments to the existing mapping rules so that similar phonetic symbols in `Pinyin' can have similar pronunciations. By combining `Pinyin' and the international standard phonetic system, the similarity among Chinese characters' pronunciations can be well described and evaluated. \\ 
In practice, we use the `pypinyin' \footnote{https://github.com/mozillazg/python-pinyin} library to acquire the `Pinyin' form of a Chinese character. Here, we will process the auxiliary, vowel, and tune separately and normally concatenate them together after being processed. 
\begin{itemize}
    \item For auxiliaries, they will be mapped to standard forms, which have at most two English characters and a phonetic weight. We apply one-hot encoding to them so that we get two one-hot vectors and a one-dimension phonetic weight. Then we add up the two English characters' one-hot vectors and the phonetic weight here will be concatenated to the tail.
    \item For vowels, they are also mapped to standard forms. However, it is a little different here. We have two different kinds of plural vowels. One is purely made up of single vowels, such as `$au$', `$eu$' and `$ai$'. The other kind is like `$an$', `$a\eta$' and `$i\eta$', which are combinations of a single vowel and a nasal voice. Here, single vowels are encoded to 6-dimension one-hot vectors and nasal voices to 2-dimension one-hot vectors respectively. Finally, We concatenate them together and if the vowel does not have a nasal voice, the last two dimensions will all be zero.
    \item  For tunes, they can be simply mapped to four-dimension one-hot vectors.
\end{itemize}

\section{Experiments}
We conduct experiments on our substitution dataset and three general datasets to verify our method from different perspectives. Standard precision, recall and, F1 scores are calculated as evaluation metrics to show the performance of different model settings. In this paper, we set up experiments on Pytorch and FastNLP \footnote{https://github.com/fastnlp/fastNLP} structure. All experiments are run on Nvidia RTX 3090 and 3080.
 
\subsection{Dataset}
\begin{table}[htb]
\begin{center}
\begin{tabular}{lrrr}
\toprule
\multirow{2}{*}{Dataset} &
\multicolumn{3}{c}{Sentences} \\
\cmidrule(r){2-4}
 & Train   & Test   & Dev \\
\midrule
Resume       & 3821   & 463      & 477     \\
Ontonote & 15724   & 4346     & 4301    \\
Weibo        & 1350   & 270      & 270     \\
Substitution & 14079   & 824      & 877     \\
\bottomrule
\end{tabular}
\caption{Dataset Composition}
\label{table1}
\end{center}
\end{table}
In order to verify whether our method has the ability to cope with the character substitution problem, we also build our own dataset. This specially designed dataset is collected from informal news reports and blogs. We label the Named Entities in raw materials first and then create their substitution forms by using similar characters to randomly replace these in the original Named Entities. In this case, the dataset is made of pairs of original entities and their character substitution forms. This dataset consists of 15780 sentences in total and is going to test our method in an extreme language environment. \\
We also conduct experiments on three general NER datasets, Chinese Resume \cite{zhang2018chinese}, Ontonote \cite{weischedel2011ontonotes}, and Weibo \cite{peng2015named,he2017f}. When introducing new features, it is important to make sure that the overall performance is not negatively affected.\\
Among three general datasets, Chinese Resume is mainly collected from resume materials. The named entities in it are mostly people's names, positions, and company names, which all have strong logic patterns. Ontonote mainly selects corpus from official news reports, whose grammar is formal and vocabulary is quite common. Different from the above three datasets, Weibo is from social media, which includes many informal expressions. \\
Details of the above four datasets are described in Table \ref{table1}.\\

\subsection{Experiment Settings}
\begin{table}[htb]
\centering
\begin{tabular}{clcrc}
\toprule
 & Items &     & Range & \\
\midrule
 & batch size  &     & 64 &     \\
 & epochs &     & 60 & \\
 & lr  &     & 2e-3  &    \\
 & optimizer &     &  Adam  & \\
 & dropout &     & 0.4  & \\
 & Early Stop  &     & 5 & \\
 & lstm layer &     & 1 & \\
\bottomrule
\end{tabular}
\caption{Hyper-parameters}
\label{table2}
\end{table}
The backbone of the NER model used in our work is mainly BiLSTM + CRF. The BiLSTM+CRF model is stable and has been verified in many research projects. Meanwhile, as our method is focused on providing a complementary lightweight module for current named entity recognition models, we select two pre-trained language models, static embedding trained by Word2Vec\cite{2013Efficient} and BERT\cite{kenton2019bert}. To be more specific, BERT-base is used in our work, which has 12 transformer layers in total. Our MFE-NER with glyph and phonetic embedding is attached to them by simple concatenation.\\
Table \ref{table2} lists some of the hyper-parameters in our training stage. Adam is used as the optimizer and the learning rate is set to 0.002. In order to reduce over-fitting, we set a rather high dropout \cite{srivastava2014dropout} rate of 0.4. Meanwhile, the Early Stop is also deployed, allowing 5 epochs of loss not decreasing. 

\section{Results and Discussions}
In this section, we first show why our method is lightweight and explain the improvement that MFE-NER achieved in recognizing substitution forms of named entities. Then, we will analyse the overall performance enhancement by applying MFE-NER. Here, the embeddings without glyph and phonetic features are named `pure' embeddings. Basically, We compare pure embeddings from pre-trained language models with those using MFE-NER for enhancement in the performance of the NER task. To control variables, we also introduce the ablation study to verify the separate contribution from glyph and phonetic features. Table \ref{table3} and \ref{table4} show all the results of different model settings in different datasets. 
\subsection{A Lightweight Method}
As a lightweight method, MFE-NER provides a convenient solution for all pre-trained models with limited extra cost. Different from other models utilizing complex techniques for data fusion, we simply alter the input layer, enlarging the dimension of character embedding by concatenation. Compared to pure semantic embeddings, MFE-NER only increases less than 5 \% of the training time and 0.4 \% of the model parameters. It is remarkable that with less than 50 dimensions added to the character embedding in the input layer, MFE-NER can cope with substitution forms while slightly improving the performance of pre-trained language models in all four datasets, showing competitive F1 scores. 
\subsection{Character Substitution Detection}
\begin{figure}[htb]
  \centering
  \includegraphics[width=0.45\textwidth]{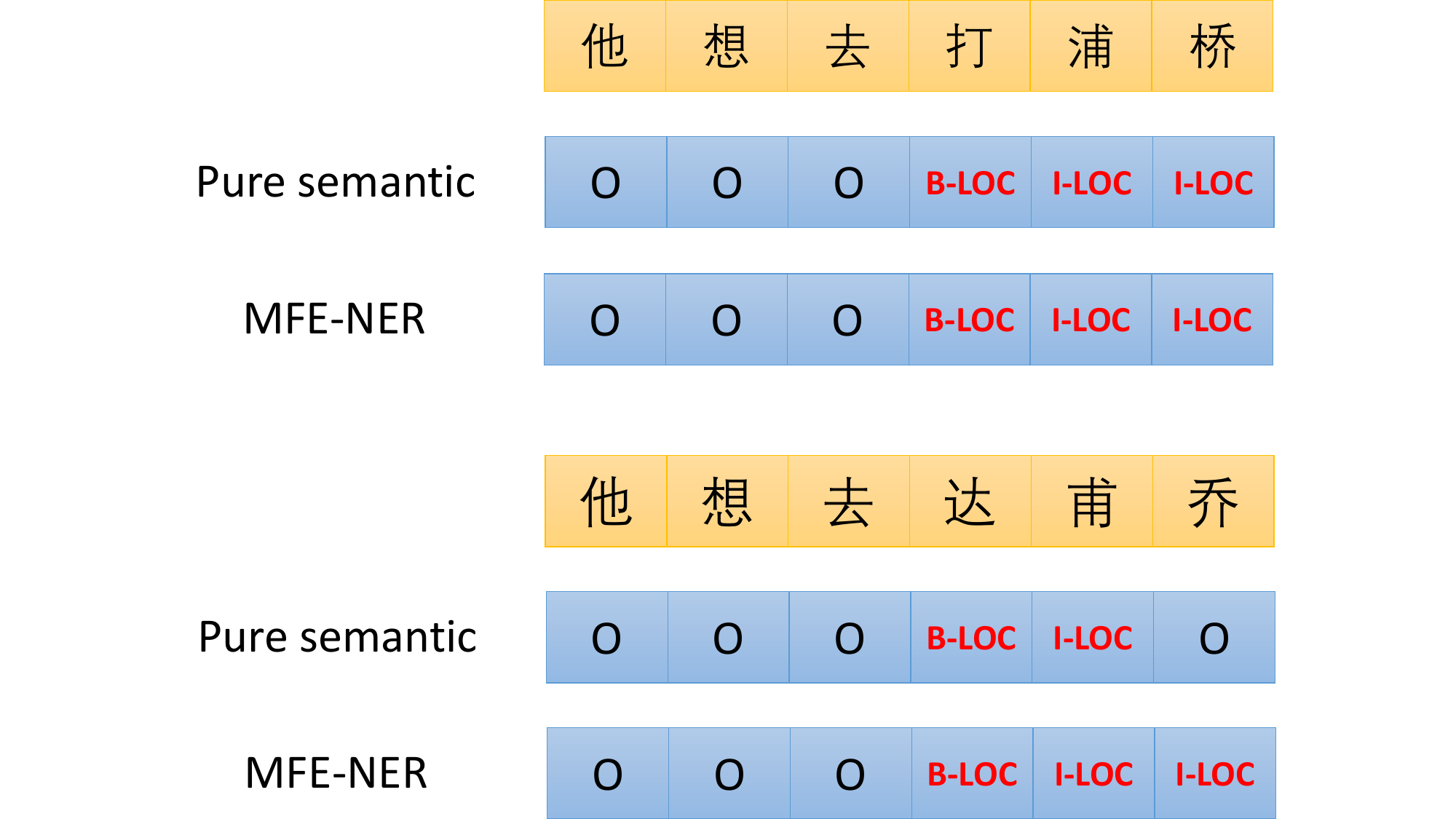}
  \caption{An example is drawn from our dataset, 'he wants to go to Dapuqiao'. The sentence at the top of the figure is the original sentence, while the sentence at the bottom is after character substitution. The model using MFE-NER gives the correct prediction.}
  \label{fig:fig5}
\end{figure}
On our Substitution dataset, MFE-NER brings remarkable advances. For different semantic embeddings, MFE-NER achieves much better performance than pure embedding, improving the F1 score by at least 1.0. \\
For the case study, there is an interesting example shown in Figure \ref{fig:fig5}. The two sentences are drawn from our dataset with the same meaning, `he wants to go to Dapuqiao'. Here, `Dapuqiao' is a famous place in Shanghai. However, the sentence below is different because characters in the original entity are changed but the word still refers to the same. Considering the model using pure semantic embedding, it fails to give the perfect prediction for the sentence below, while the model using MFE-NER can exploit the extra information, thus giving the perfect prediction.\\
MFE-NER also shows its advantages in some informal language environments. The Weibo dataset is a typical example from social media. On the Weibo dataset, MFE-NER achieves 53.81 and 67.74 in models with static embedding and BERT respectively, significantly enhancing the performance of pre-trained language models. \\
\subsection{Overall NER Performance}
\begin{table*}[t]
\centering
\begin{tabular}{ccccccc}
\toprule
\multirow{2}{*}{Models} &
\multicolumn{3}{c}{Weibo} & \multicolumn{3}{c}{Substitution} \\
\cmidrule(r){2-4} \cmidrule(r){5-7}  & Precision   & Recall   & F1 & Precision   & Recall   & F1 \\
\midrule
pure Word2Vec & 67.19 & 40.67 & 50.67 & \textbf{84.60} & 71.24 & 77.35\\
glyph Word2Vec  & 62.54 & 43.54 & 51.34 & 80.80 & 74.66 & 77.61\\
phonetic Word2Vec  & 66.06 & 43.30 & 52.31 & 84.09 & 74.81 & 79.18\\
MFE-NER (Word2Vec) & \textbf{67.51} & \textbf{44.74} & \textbf{53.81} & 84.36 & \textbf{75.56} & \textbf{79.72}\\
\midrule
pure BERT   & 68.48 & 63.40 & 65.84 & 88.67 & 81.67 & 85.03\\
glyph BERT   & 71.88 & 64.83 & 68.18 & 89.39 & \textbf{82.86} & 86.01\\
phonetic BERT   & 70.30 & \textbf{66.27} & \textbf{68.22} & \textbf{90.48} & 82.12 & 86.09\\
MFE-NER (BERT)  & \textbf{73.54} & 63.16 & 67.95 & 90.08   & 82.56     & \textbf{86.16}\\
\bottomrule
\end{tabular}
\caption{Results on two datasets in informal language environments, Weibo, and Substitution.}
\label{table3}
\end{table*}

\begin{table*}[t]
\centering
\begin{tabular}{ccccccc}
\toprule
\multirow{2}{*}{Models} &
\multicolumn{3}{c}{Chinese Resume} & \multicolumn{3}{c}{Ontonotes} \\
\cmidrule(r){2-4} \cmidrule(r){5-7} & Precision   & Recall   & F1 & Precision   & Recall   & F1 \\
\midrule
pure Word2Vec  & 93.97 & 93.62 & 93.79 & 72.92 & 57.83 & 64.51 \\
glyph Word2Vec & 93.87 & 93.93 & 93.90 & 71.34 & 59.88 & \textbf{65.11} \\
phonetic Word2Vec  & 93.58 & 93.93 & 93.75 & \textbf{73.39} & 57.12 & 64.24 \\
MFE-NER (Word2Vec) & \textbf{94.29} & \textbf{94.17} & \textbf{94.23} & 70.92 & \textbf{59.88} & 64.93 \\
\midrule
pure BERT   & 94.60 & 95.64 & 95.12 & 80.53 & 80.93 & 80.73 \\
glyph BERT   & 94.83 & 95.58 & 95.20 & \textbf{83.44} & 79.49 & \textbf{81.42} \\
phonetic BERT   & 95.47 & \textbf{95.77} & 95.62 & 81.30 & 79.85 & 80.57 \\
MFE-NER (BERT)  & \textbf{95.76} & 95.71 & \textbf{95.73} & 80.61 & \textbf{81.87} & 81.24 \\
\bottomrule
\end{tabular}
\caption{Results on the datasets in formal language environments, Chinese Resume and Ontonotes}
\label{table4}
\end{table*}
Experiments on the other two general datasets from formal language environments also show that MFE-NER brings slight improvement to the overall NER task. On Chinese Resume, MFE-NER with BERT achieves a 95.73 F1 score and MFE-NER with static embedding gets a 94.23 F1 score, improving the performance of pure semantic embedding from pre-trained language models by about 0.5 with respect to F1 score. On Ontonote, MFE-NER also increases the F1 score of static embedding from 80.73 to 81.24 and boosts the performance of using BERT as semantic embedding. However, owing to fewer grammar mistakes and substitution forms in the standard language environment, the performance of MFE-NER is limited.\\
Slight improvement still makes sense. Considering that, in Chinese, named entities have their own glyph and phonetic patterns. For example, in the glyph domain, characters in Chinese names usually contain a special character root, which denotes `people'. What's more, characters representing places and objects also include certain character roots, which show materials like water, wood, and soil. These character roots can be utilized in `Five-Strokes', thus improving the performance of overall NER performance.\\
Based on the experiment results above, MFE-NER is able to reduce the negative influence of the character substitution phenomenon in Chinese Named Entity Recognition, while slightly improving the overall performance of NER models. It makes sense that MFE-NER is suitable to solve character substitution problems because glyph and phonetic features are introduced, which provide more prior knowledge. These extra features are complementary to semantic embedding from pre-trained models and bring information that offers more concrete proofs to NER models.\\

\subsection{Ablation Study}
Ablation study is thus made to investigate how glyph and phonetic features bring improvement by themselves. We separately add glyph embedding or phonetic embedding to pre-trained language models for comparisons. From the results on all four datasets, it is clear that models with the glyph or phonetic embedding almost all perform better than models with pure semantic embedding, which means that extra patterns from glyph and phonetic domains are all helpful in the NER task, strengthening the original model ability.\\
However, in some datasets, the models adding glyph or phonetic embedding only to BERT achieve higher F1 scores than MFE-NER. The reasons are to follow. On one hand, named entities in different datasets may rely on one of our provided features much more than the other features. So, in the test stage, this specific feature would contribute more to identifying named entities in the corpus. On the other hand, when we add extra information to the existing pre-trained language models, considering the different data distributions of train, dev, and test set, the model we select might give higher weights to the learned patterns from glyph or phonetic domain, which shows good performance on dev set but possibly leads to over-fitting.

\section{Conclusion}
Nowadays, the informal language environment created by social media has deeply changed the way that people express their thoughts. Using character substitution to generate new named entities becomes a common linguistic phenomenon which is a big challenge for NER. In this paper, we propose a lightweight method fusing the glyph and phonetic features 
for Chinese Named Entity Recognition (MFE-NER) to handle the Chinese character substitution problem, which introduces 'Five-strokes' to represent glyph features, making it more convenient to differentiate similar characters, and develops a system named `Trans-Pinyin' to represent phonetic features of Chinese characters, solving the problem of similar pronunciations in Pinyin. Experiments show that our model, MFE-NER, can assist pre-trained models and reduce the influence caused by character substitution. \\

\section*{Acknowledgments}
We thank anonymous reviewers for their helpful comments. 

\bibliography{ijcai22}
\bibliographystyle{ccl.bst}

\end{document}